\documentclass{article}

\usepackage[nonatbib,preprint]{neurips_2025}
\usepackage[numbers]{natbib}

\usepackage{dsfont}
\usepackage{indentfirst}
 \usepackage{amsfonts}

\usepackage{graphicx}
\usepackage{amsmath}

\usepackage{amssymb}
\usepackage{booktabs}

\usepackage{float}
\usepackage{multirow}
\usepackage{diagbox}
\usepackage{balance}
\usepackage[linesnumbered,titlenumbered,ruled,vlined,commentsnumbered,noend]{algorithm2e}

\usepackage[T1]{fontenc}
\usepackage{xcolor}
\usepackage{epsfig}

\usepackage{enumitem}
\setenumerate[1]{itemsep=0pt,partopsep=0pt,parsep=\parskip,topsep=5pt}
\setitemize[1]{itemsep=0pt,partopsep=0pt,parsep=\parskip,topsep=5pt}
\setdescription{itemsep=0pt,partopsep=0pt,parsep=\parskip,topsep=5pt}

\usepackage{moresize}
\usepackage{epstopdf}
\usepackage{array}

\usepackage{makecell}
\usepackage{bm}
\usepackage{dsfont}

\usepackage{amssymb}
\usepackage{sidecap}
\usepackage{colortbl}

\usepackage[font=footnotesize,labelfont=bf]{caption}
\definecolor{citecolor}{RGB}{65,105,225}
\usepackage[breaklinks=true,colorlinks,citecolor=citecolor,bookmarks=false]{hyperref}

\definecolor{tab_red}{RGB}{255,174, 185}
\definecolor{tab_yellow}{rgb}{0.99,0.99,0.70}

\usepackage{dsfont}
\usepackage{xspace}

\def\argmin{\operatornamewithlimits{arg\,min}}

\definecolor{dg}{rgb}{0,0.694,0.298}
\definecolor{purple}{rgb}{0.4,0.176,0.569}
\definecolor{royalblue}{RGB}{65,105,225}

\usepackage{pifont}
\usepackage{xspace}
\makeatletter
\DeclareRobustCommand\onedot{\futurelet\@let@token\@onedot}
\def\@onedot{\ifx\@let@token.\else.\null\fi\xspace}
 
\def\ie{\emph{i.e}\onedot}

\makeatother

\definecolor{americanrose}{rgb}{1.0, 0.01, 0.24}

\usepackage[capitalize]{cleveref}
\crefname{section}{Sec.}{Secs.}
\Crefname{section}{Section}{Sections}
\Crefname{table}{Table}{Tables}
\crefname{table}{Tab.}{Tabs.}

\usepackage[utf8]{inputenc} 
\usepackage[T1]{fontenc}    
\usepackage{hyperref}       
\usepackage{url}            
\usepackage{booktabs}       
\usepackage{amsfonts}       
\usepackage{nicefrac}       
\usepackage{microtype}      
\usepackage{xcolor}         
\usepackage{wrapfig}

\title{Concept Guided Co-salient Object Detection}

\author{Jiayi Zhu\textsuperscript{1}, Qing Guo\textsuperscript{3}, Felix Juefei-Xu\textsuperscript{2}, Yihao Huang\textsuperscript{4}, Yang Liu\textsuperscript{4}, Geguang Pu\textsuperscript{1,5}\\
~\\
\textsuperscript{1} East China Normal University, China \quad
\textsuperscript{2} New York University, USA \\
\textsuperscript{3} IHPC \& CFAR, Agency for Science, Technology and Research, Singapore \\
\textsuperscript{4} Nanyang Technological University, Singapore \\
\textsuperscript{5} Shanghai Industrial Control Safety Innovation Tech. Co., Ltd, China
}

\begin{document}

\maketitle
\begin{abstract}
\indent
Co-salient object detection (Co-SOD) aims to identify common salient objects across a group of related images.  While recent methods have made notable progress, they typically rely on low-level visual patterns and lack semantic priors, limiting their detection performance. We propose ConceptCoSOD, a concept-guided framework that introduces high-level semantic knowledge to enhance co-saliency detection. By extracting shared text-based concepts from the input image group, ConceptCoSOD provides semantic guidance that anchors the detection process. To further improve concept quality, we analyze the effect of diffusion timesteps and design a resampling strategy that selects more informative steps for learning robust concepts.  This semantic prior, combined with the resampling-enhanced representation, enables accurate and consistent segmentation even in challenging visual conditions.  Extensive experiments on three benchmark datasets and five corrupted settings demonstrate that ConceptCoSOD significantly outperforms existing methods in both accuracy and generalization.
\end{abstract}
\section{Introduction}
\label{sec:intro}
Co-salient object detection (Co-SOD) plays a crucial role in numerous computer vision applications, including data preprocessing for weakly supervised semantic segmentation \cite{zeng2019joint} and video object localization \cite{jerripothula2016cats,jerripothula2018efficient}. The goal of Co-SOD is to detect common and prominent objects by analyzing the intrinsic relationships within related image groups. This task demands the simultaneous processing of multiple images to extract and align shared features, presenting considerable challenges due to the substantial variation in object appearance, orientation, and background across different scenes. 


Previous Co-SOD research explored various approaches. Early methods \cite{chang2011co,fu2013cluster,li2011co} relied on handcrafted features like color and texture but lacked robustness and scalability. More recent learning-based techniques, such as deep learning \cite{li2019detecting,wei2019deep}, self-paced learning \cite{hsu2018unsupervised,zhang2016co}, metric learning \cite{han2017unified}, and graph reasoning \cite{jiang2019unified,zheng2018feature}, have significantly improved semantic modeling and overall performance. 
However, despite these advances, most existing methods overlook the crucial role of semantic priors, revealing a fundamental flaw in their design. They typically approach object discovery as a low-level pattern matching task, relying on visual cues such as color, texture, or spatial layout, without incorporating high-level knowledge about object categories or concepts. Without such priors, the model lacks a conceptual anchor to distinguish essential features from incidental ones. As a result, it struggles with \textit{inter-object variations} because appearance differences, such as changes in pose or scale, can obscure the underlying semantic identity of the object. Similarly, the absence of semantic grounding makes the model vulnerable to \textit{background distractions}, where non-target elements that share similar visual patterns with the object can easily mislead detection. In both cases, the model fails not because of insufficient data or inadequate architecture, but because it lacks a high-level understanding of what to look for.

To address these issues, we propose incorporating semantic priors into the Co-SOD pipeline. Specifically, we introduce a concept-guided framework that leverages the rich object-level knowledge encoded in vision-language models such as diffusion-based text-to-image models. Through a personalization process, we extract a concept representation of the target object that captures high-level semantics and structural cues. This representation serves as a semantic anchor that guides the model in identifying the true object of interest, thereby improving robustness to both inter-object variations and background distractions. Our approach consists of two key modules: \ding{182} Given a group of related images, we first extract text-based semantic information associated with the target object, forming a personalized concept that reflects real-world diversity in pose, appearance, and perspective. \ding{183} This concept is then used to guide fine-grained segmentation of the target object across the image group. To further improve task performance, we analyze the effect of the concept-learning process on Co-SOD outcomes and identify that the choice of timestep in the diffusion model significantly impacts final detection quality. Based on this observation, we design a novel timestep resampling strategy for more effective object localization.  This leads to a clear improvement in overall Co-SOD accuracy and robustness. We term the overall end-to-end framework \textit{ConceptCoSOD}.

Our main contributions are summarized as follows:
\begin{itemize}[itemsep=2pt, topsep=0pt, parsep=0pt, left=10pt]
\item We propose the first concept-guided Co-SOD framework that explicitly integrates semantic priors from diffusion models. Unlike previous methods based on low-level pattern matching, our approach uses high-level concepts from the image group to guide co-saliency detection, offering a more principled and semantically informed solution.
\item We propose a two-stage pipeline: a concept learning module to distill object-level semantics and a concept-guided segmentation module for precise object localization.
\item We further introduce a timestep resampling strategy, motivated by an analysis of diffusion timesteps, which enhances concept quality and further boosts Co-SOD performance.
\item Extensive experiments on seven datasets and five baselines demonstrate the effectiveness and robustness of the proposed \textit{ConceptCoSOD} method.
\end{itemize}


\section{Related Work}
\label{sec:related}

\subsection{Co-salient Object Detection}

Co-Saliency Detection (Co-SOD) identifies shared salient objects in related images by exploiting their intrinsic relationships \cite{tang2024co,zhang2021deepacg}. Early methods \cite{li2013co,song2016rgbd,jerripothula2016cats} used handcrafted features, which lacked robustness in complex scenes. Recent deep learning approaches \cite{fan2021group,zhang2020gradient} improve performance by forming a group-wise consensus representation and redistributing it to refine image saliency.

For instance, Wang et al. \cite{wang2015saliency} used summation-based feature aggregation with gradient feedback. Jin et al. \cite{jin2020icnet} proposed ICNet, incorporating enhanced intra-saliency vectors and dense correlations. Zhang et al. \cite{zhang2020gradient} introduced GICD, emphasizing discriminative kernels via gradient induction. Zhang et al. \cite{zhang2020adaptive} developed GCAGC, an adaptive graph convolutional network with attention-based clustering. Fan et al. \cite{fan2021group} built GCoNet using an affinity-based consensus attention map. CADC \cite{zhang2021summarize} used dynamic kernels for effective consensus distribution. Lastly, Yu et al. \cite{yu2022democracy} presented DCFM, a democratic feature mining framework enhancing detection without extra inputs.

\subsection{Text-to-Image Diffusion Generation Model}

Text-to-image (T2I) generation \cite{zhang2023text}, popularized by diffusion models \cite{croitoru2023diffusion,ho2020denoising,rombach2022high}, excels at producing diverse, realistic images from text prompts. This began with Ho et al.’s denoising diffusion probabilistic models (DDPMs) \cite{ho2020denoising}, which rivaled PGGAN \cite{karras2018progressive}. Later, the latent diffusion model (LDM) \cite{rombach2022high} was proposed to perform denoising in latent space, followed by decoding into detailed images, thereby significantly enhancing generation speed and reducing computational complexity.
Stable Diffusion \cite{stable_diffusion}, a widely used T2I model, is based on LDM.

\subsection{Concept Learning}
In concept learning (personalization), users provide a few example images to generate new scenes featuring the desired concept via text prompts. 
Current personalization methods typically follow one of two approaches: encapsulating a concept via a word embedding at the input of the text encoder \cite{gal2023an,daras2022multiresolution}, or fine-tuning the weights of diffusion-based modules through various techniques \cite{ruiz2022dreambooth,hu2021lora,gal2023designing,shi2023instantbooth}. 
Our approach utilizes these personalization techniques to extract shared semantics across image groups.


\section{Preliminaries}
\label{sec:preliminary}
\subsection{Problem Formulation}
Given a group of images $\mathcal{I} = \{\mathbf{I}_i \in \mathds{R}^{H\times W\times 3}\}_{i=1}^{N}$ containing $N$ images that have common salient objects, the co-saliency object detection (Co-SOD) task is to detect and segment all these objects in the image group.
We denote a Co-SOD method as $\textsc{CoSOD}(\cdot)$ and predict $N$ salient maps via
\begin{align}
    \mathcal{S} = \{\mathbf{S}_i\}_{i=1}^{N} = \textsc{CoSOD}(\mathcal{I})    \label{eq:cosal},
\end{align}
where $\mathbf{S}_i\in \mathds{R}^{H\times W}$ is a binary saliency map corresponding to the salient region of $\mathbf{I}_i$.

\subsection{Motivation}


We investigated existing Co-SOD methods and revealed that they struggle to maintain performance when dealing with complex scenes. The reason is that these Co-SOD methods typically treat the discovery of co-salient objects as a low-level pattern matching task, relying heavily on visual cues such as color, texture, or spatial layout. Although these methods are capable of capturing surface-level similarities among objects, they often lack the ability to model deeper semantic relationships essential for the co-salient object understanding.

\begin{figure}[]
	\centering   
	\includegraphics[width=1\columnwidth]{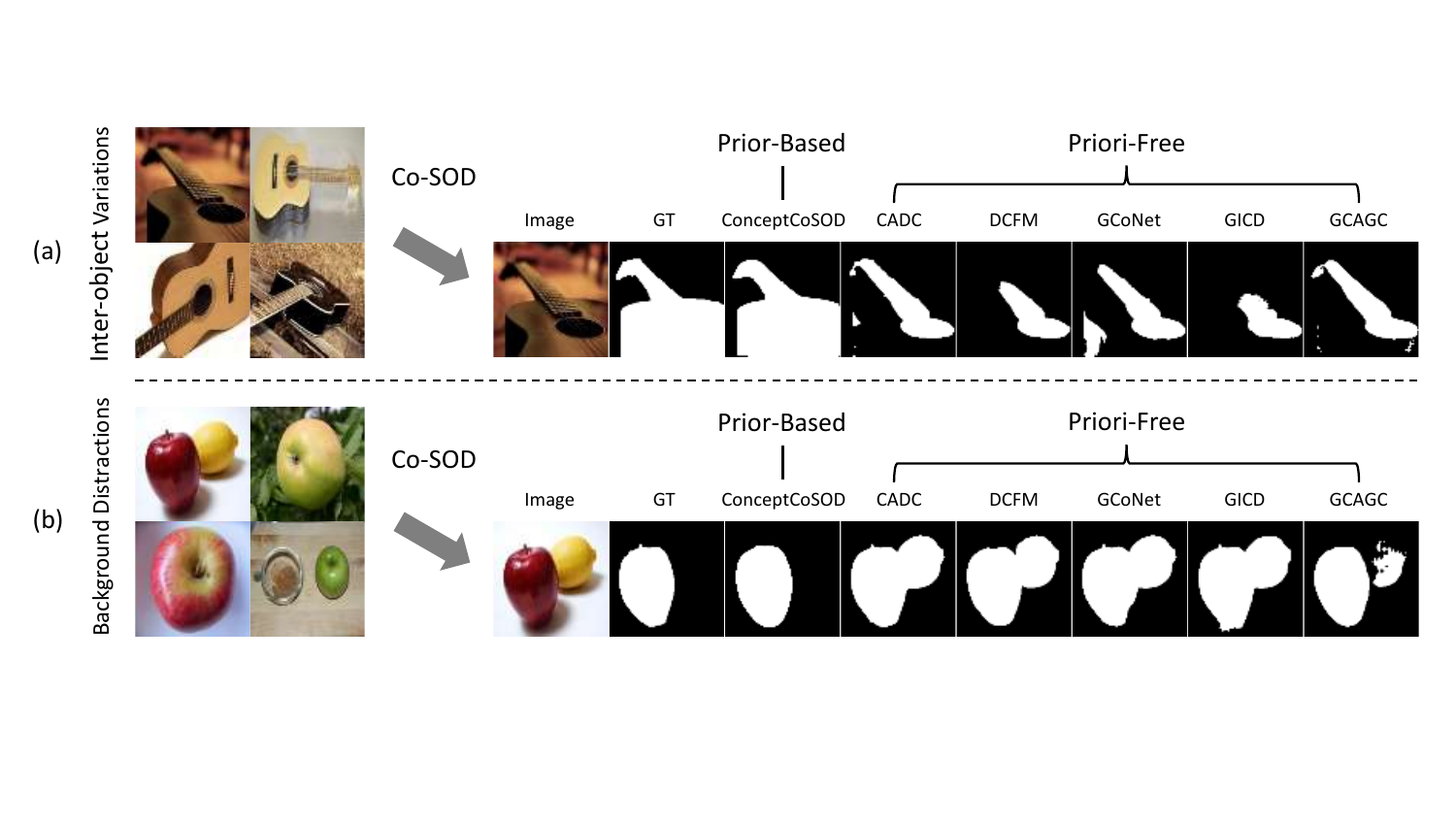}
	\caption{Limitations of existing prior-free Co-SOD methods.}
	\label{fig:motivation}
\end{figure}

As illustrated in Fig.\ref{fig:motivation}, existing Co-SOD methods exhibit significant inaccuracies in both presented examples. In Fig.\ref{fig:motivation} (a), the group images contain guitars viewed from different angles. The first image presents a bottom-up perspective, where current state-of-the-art (SOTA) methods consistently fail to detect the entire guitar. Instead, they isolate only the guitar strings and erroneously treat them as the complete object. This error stems from the lack of prior knowledge about the object’s structural composition. Without understanding that a guitar is made up of components such as the body, neck, and headstock, the model misinterprets a partial observation as the whole. In Fig.~\ref{fig:motivation} (b), the co-salient object is an apple, yet the first image also contains a lemon. Without any conceptual understanding of what constitutes an apple, the model fails to recognize the lemon as a separate entity and mistakenly treats it as part of the apple. This further reflects the absence of high-level semantic priors that are necessary for distinguishing object boundaries and identities. These examples reveal a fundamental limitation of prior-free Co-SOD methods. Lacking semantic understanding, these models rely solely on low-level appearance consistency across images and fail to reason about object completeness or category membership. Consequently, they misidentify parts as wholes and incorporate unrelated distractors into the co-salient target, especially under viewpoint variation or background clutter.

Therefore, it is necessary to incorporate prior knowledge into the Co-SOD process.
Existing vision-language models have been extensively trained and possess rich semantic knowledge. We aim to leverage their capabilities by incorporating this prior knowledge into our Co-SOD framework.
We drew inspiration from the effective practice of vision-language methods in segmentation tasks \cite{barsellotti2024training,xu2023open,li2023open,wu2023diffumask}, which exploit text semantics as valuable supplementary information.

Although leveraging text semantics for the Co-SOD task is promising, no textual information is directly available within the task. Hence, we propose to extract effective semantic information directly from the group of images first and then utilize it to guide subsequent detection. Thus, we reformulate the method from Eq.~\eqref{eq:cosal} to a new one as follows:
\begin{align}
    \mathcal{S} = \{\mathbf{S}_i\}_{i=1}^{N} = \textsc{CoSOD}(\mathcal{I},\text{Extract}(\mathcal{I}))  \label{eq:cosal_reformulate},
\end{align}
where $\text{Extract}(\cdot)$ denotes the text semantic extraction.

%

%








\section{Methodology: \textsc{ConceptCoSOD}}
\label{sec:method}

\subsection{Overview}\label{subsec:overview}
We propose a method termed \textit{ConceptCoSOD}, which comprises two modules.
The first module $\text{ConceptLearn}(\cdot)$ involves utilizing a group of input images for learning the text embedding (\ie, concept $\mathbf{c}^*$) of the co-existing object in the image group $\mathcal{I}$:
\begin{align} \label{eq:conceptlearn_ConceptCoSOD}
    \mathbf{c}^* = \text{ConceptLearn}(\mathcal{I}),
\end{align}

The second module $\text{SOD}(\cdot)$ uses the learned concept as guidance to generate co-saliency object detection results through an end-to-end approach. Specifically, the learned concept is applied to perform object detection on each image $\mathbf{I}$ in the image group $\mathcal{I}$:
\begin{align}
	\label{eq:each_SOD}
    {\mathbf{S}_i} = \text{SOD}(\mathbf{c}^*,\mathbf{I}_i), \mathbf{I}_i\in \mathcal{I}.
\end{align}
The detailed explanation of the two modules is demonstrated in Sec.~\ref{subsec:concept_learn} and Sec.~\ref{subsec:concept_guided_seg}.


\subsection{Concept Learning Module}\label{subsec:concept_learn}

\begin{figure}
\centering   
\includegraphics[width=\linewidth]
    {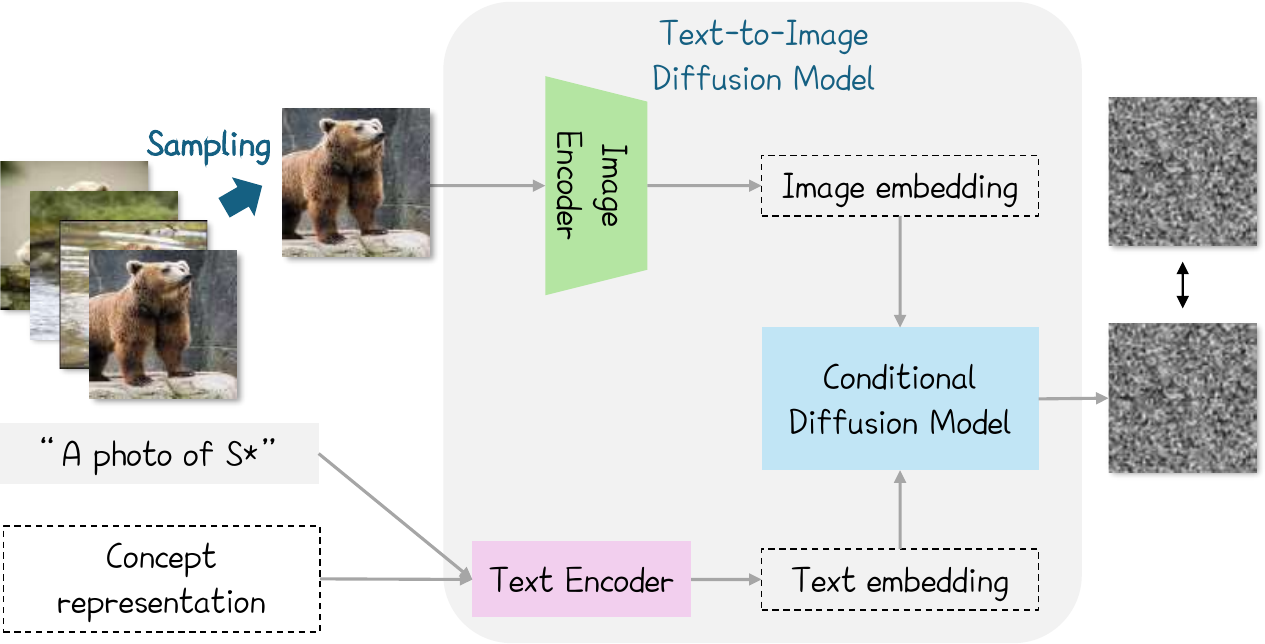}
	\caption{Group-Image concept learning module.}
	\label{fig:concept_extraction}
\end{figure}

This module learns the semantic concept $\mathbf{c}^*$ from the image group $\mathcal{I}$, inspired by personalization methods \cite{gal2023an} that extract object-level concepts from a few images. Specifically, $\mathbf{c}^*$ is a learnable token in the latent space of a text-to-image diffusion model, representing the shared object across the group. The module architecture is illustrated in Fig.~\ref{fig:concept_extraction}.

We adopt a text-to-image diffusion model \cite{rombach2022high} to guide concept learning, consisting of three components: (1) an image autoencoder with encoder $\mathcal{E}$, and decoder $\mathcal{D}$, where $\mathcal{E}$ maps an image $\mathbf{X}$ to a latent vector $\mathbf{z} = \mathcal{E}(\mathbf{X})$ and $\mathcal{D}$ can reconstruct it with $\mathcal{D}(\mathcal{E}(\mathbf{X})) \approx \mathbf{X}$, (2) a text encoder $\Gamma$ that converts a text prompt $\mathbf{y}$ into a latent embedding $\Gamma(\mathbf{y})$, and (3) a conditional diffusion model $\epsilon_\theta$ that can predict the noise added to $\textbf{z}_t$ at time step $t$, denoted as $\epsilon_\theta(\textbf{z}_t,t,\Gamma(\textbf{y}))$, with time step $t$, the noise latent variable at step $t$, $\textbf{z}_t$, and the text representation $\Gamma(\textbf{y})$ as input.

Given the image group $\mathcal{I}$ and the pretrained text-to-image diffusion model, we aspire to learn a concept of $\mathcal{I}$ via the following optimization:
\begin{align}
    %
    \textbf{c}^* = & \argmin\limits _{\textbf{c}} \mathbb{E}_{\textbf{z}, \mathbf{y},\epsilon\in \mathcal{N}(0,1),t \sim \mathcal{U}([0, 1])} (\|\epsilon_\theta(\mathbf{z}_t,t, \Phi(\Gamma(\mathbf{y}),\textbf{c}))-\epsilon\|_2^2),
    \label{eq:argmin_concept_ConceptCoSOD}
\end{align}
where $\mathbf{y}$ is a fixed text prompt (e.g., `a photo of $S^*$'), and $\Phi(\Gamma(\mathbf{y}), \mathbf{c})$ denotes the injection process that replaces the `$S^*$' token in $\Gamma(\mathbf{y})$ with the learnable embedding $\mathbf{c}$.
Eq.~\eqref{eq:argmin_concept_ConceptCoSOD} encourages the optimized concept $\mathbf{c}^*$ to encode the co-salient object positions in the text latent space.
Once $\mathbf{c}^*$ is obtained, it can be inserted into other prompts to guide new image generation via a text-to-image diffusion model.


\subsection{Concept-Guided Segmentation}
\label{subsec:concept_guided_seg}

\begin{figure}
\centering   
\includegraphics[width=\linewidth]{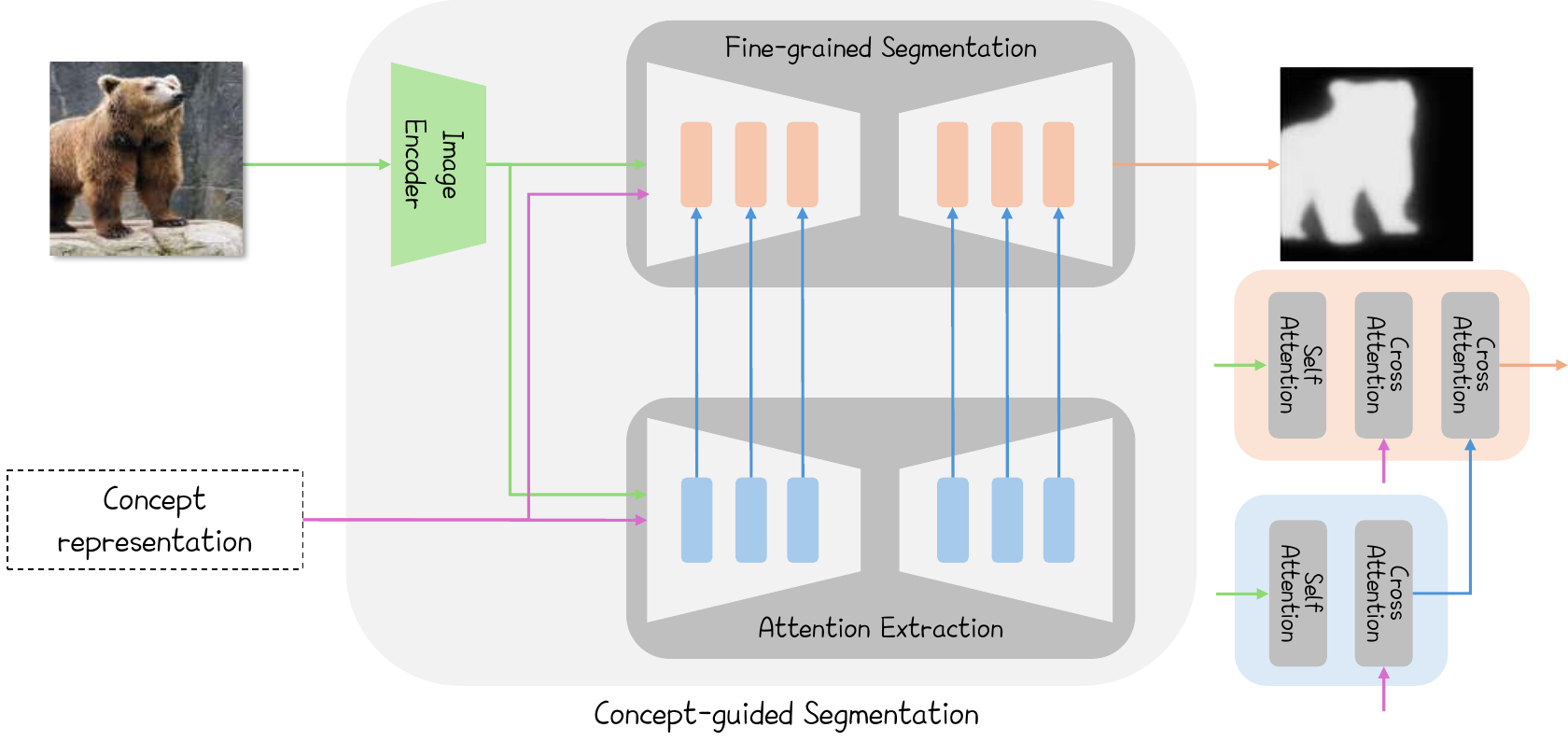}
	\caption{Concept-guided segmentation module.}
	\label{fig:concept_guided_segmentation}
\end{figure}

The concept-guided segmentation module contains two submodules. As shown in Fig.~\ref{fig:concept_guided_segmentation}, we use an attention extraction submodule to extract the coarse attention map first and then take it as the guidance input for the fine-grained segmentation submodule to output the final segmentation result. Both the submodules are classical U-Net architectures used in text-to-image models whose process takes multi-modality input (refer to architecture \cite{rombach2022high}).

Specifically, given an image $\mathbf{I}$ and its corresponding concept representation $\textbf{c}^*$, the concept-guided segmentation module first transforms the image into a latent representation $\mathbf{z}$ at time step $t$. Using this latent $\mathbf{z}_t$ and concept representation $\textbf{c}^*$ as input, the attention extraction submodule generates an attention map following $\textbf{f}_{attn} = \text{AttnExtract}(\epsilon_\theta,\mathbf{z}_t,t,\textbf{c}^*)$. Notably, the fine-grained segmentation submodule differs slightly from the traditional U-Net in text-to-image model by incorporating an additional cross-attention mechanism to process the attention map $\textbf{f}_{attn}$ as supplementary guidance alongside the prior guidance (i.e., concept representation $\textbf{c}^*$). With latent $\mathbf{z}_t$, concept representation $\textbf{c}^*$, and attention map $\textbf{f}_{attn}$ as inputs, the fine-grained segmentation submodule (i.e., function $\text{FineSeg}(\cdot)$) leverages these components to produce the final segmentation output,
\begin{align}
    \mathbf{S}' = \text{FineSeg}(\epsilon_\phi, \mathbf{z}_t,t,\textbf{c}^*,\textbf{f}_{attn}),
    \label{eq:znet}
\end{align}
where $\epsilon_\phi$ denotes the parameters in the module. Finally, we apply a predefined threshold value $\lambda$ to $\mathbf{S}'$ for final processing. In subsequent experiments, we discuss the impact of choosing different threshold values $\lambda$ on the co-saliency object detection results. For each pixel $s$ in $\mathbf{S}'$, the value satisfies:
\begin{align}
\label{threshold}
	\mathbf{S} = \begin{cases}
	1, & \text{if } s \geq \lambda \\
	0, & \text{otherwise}
	\end{cases}
\end{align}

\subsection{Timestep Resampling Strategy}
\label{subsec:timestep_resampling}
\begin{figure}[tbp]
	\centering   
	\includegraphics[width=1\columnwidth]{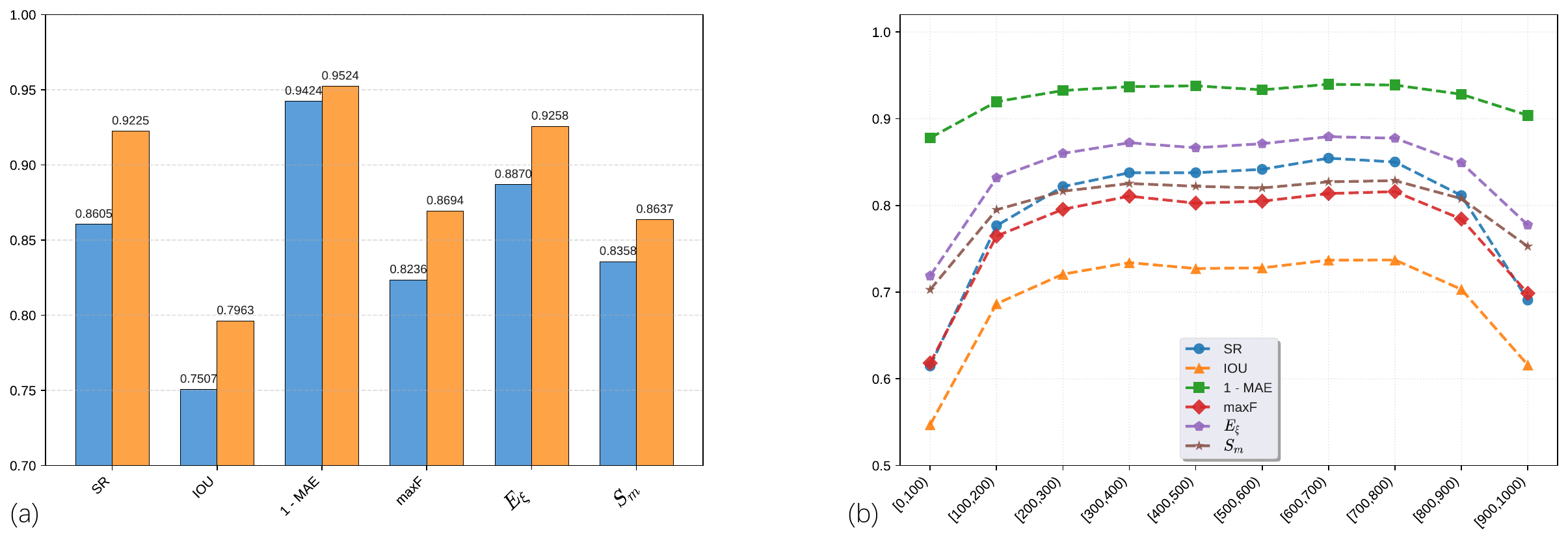}
	\caption{
    Exploring the potential of ConceptCoSOD and the effectiveness of different timestep ranges. \\(a) Comparison between the base version of ConceptCoSOD (\ie, without resampling strategy) and the theoretical upper bound (\ie, with accurate tokens). (b) Performance comparison across ten different timestep ranges.
	}
	\label{fig:strategy_compare}
\end{figure}

It is evident that the quality of the concept learned from input images plays a critical role in determining segmentation performance in the CoSOD task. This raises an important question: is the concept representation learned by existing personalization methods sufficiently effective for CoSOD? To investigate this, we compare the segmentation performance achieved using the personalized concept representation $\textbf{c}^*$ with that obtained using native token representations (e.g., the textual tokens ``cat'', ``dog'') directly from the token dictionary of a text-to-image model. As shown in Figure~\ref{fig:strategy_compare} (a), native token-based concepts achieve significantly better segmentation performance than the personalized concept, suggesting that the concept learned through personalization still leaves considerable room for improvement in the context of the CoSOD task.

To improve concept learning from input images in the CoSOD task, we conduct an empirical study on a critical factor in diffusion models: the sampling timestep. As shown in Figure~\ref{fig:strategy_compare} (b), our results demonstrate that the choice of timestep significantly affects the final segmentation performance. Motivated by this observation, we design a timestep resampling strategy to enhance concept learning. Specifically, based on the performance trends in Figure~\ref{fig:strategy_compare} (b), we introduce a resampling method controlled by a ratio parameter $\alpha$. We divide the full diffusion range $T$ into three intervals: the head $T_1$, the middle $T_2$, and the tail $T_3$.
The resampling distribution $p_{\alpha}(t)$ is defined as:
$$
p_{\alpha}(t) = 
\begin{cases}
\frac{1}{2 \cdot |T_1|\cdot(1+\alpha)}, & t \in T_1 \\
\frac{\alpha}{|T_2|\cdot(1+\alpha)}, & t \in T_2 \\
\frac{1}{2 \cdot |T_3|\cdot(1+\alpha)}, & t \in T_3 \\
\end{cases}
$$
This formulation ensures that a higher value of $\alpha$ increases the sampling probability in the informative middle interval $T_2$, while proportionally decreasing it in the less effective regions $T_1$ and $T_3$. The distribution is normalized so that the total sampling mass remains one, thereby preserving the overall sampling budget.

\section{Experiment}
\label{sec:experiment}

\subsection{Experimental Setup} 
\noindent\textbf{Datasets.}
Our experiments on the clean datasets were conducted on Cosal2015 \cite{zhang2016detection}, CoSOD3k \cite{fan2020taking}, and CoCA \cite{zhao2019egnet}.
These three datasets contain 2015, 3316, and 1295 images of 50, 160, and 80 groups respectively.
In terms of the corrupted datasets, we applied five types of corruption methods to the Cosal2015 dataset, specifically including frost, motion blur, defocus blur, and Gaussian noise from common corruption types \cite{hendrycks2019robustness}, as well as the adversarial attack Jadena \cite{gao2022can}.
Specifically, we applied corruption to the first $50\%$ of images in each image group while keeping the remaining $50\%$ of images clean.


\noindent\textbf{Baseline methods.}
We selected five comparison baseline methods that have demonstrated excellent performance in co-saliency object detection, including GCAGC \cite{zhang2020adaptive}, GICD \cite{zhang2020gradient}, GCoNet \cite{fan2021group}, DCFM \cite{yu2022democracy} and CADC \cite{zhang2021summarize}.

\noindent\textbf{Metrics.} 
We employ six metrics to evaluate co-salient object detection performance. Intersection over Union (IoU) assesses the overlap between the detected region and the ground truth, with the average IoU calculated across all detections. Detections with an IoU value above 0.5 are considered successful, and the success rate (SR) serves as an additional metric. Mean Absolute Error (MAE) \cite{zhang2018review} quantifies the average pixel-wise difference between the Co-SOD result and ground truth. Maximum F-measure (maxF) \cite{achanta2009frequency} calculates a harmonic mean of precision and recall, balancing these aspects in Co-SOD evaluations. The Enhanced Alignment Measure ($E_\xi$) \cite{fan2018enhanced} captures alignment and distribution by integrating pixel-level and image-level assessments. Lastly, the Structure Measure ($S_m$) \cite{fan2017structure} evaluates structural similarities, considering both region-based and object-based coherence.

\noindent\textbf{Implementation details.} 
In the group-image concept learning module, we adopt the Textual Inversion \cite{gal2022image} and employ Stable Diffusion v1.4 in our implementation.
Since the overall timestep range is $[0,1000)$, the head interval $T_1$, the middle interval $T_2$, and the tail interval $T_3$ are set to $[0,300)$, $[300,800)$ and $[800,1000)$, respectively.
The ratio parameter $\alpha$ for timestep resampling is set to 1.5.
The learning rate for updating $\textbf{c}^*$ is set to 5$e^{-4}$, the learning batch size is set to 4 and the max learn step is set to 2,000.
%
%
In the concept-guided segmentation module, we employ an existing segmentation method LD-ZNet \cite{pnvr2023ld} and use the concept we learned for guidance.
The threshold for segmentation is set to 0.5.
Experiments are run on an Ubuntu system with an NVIDIA RTX A6000 with 48G RAM.

\subsection{Comparison on Clean Datasets}
We compare our ConceptCoSOD method with baselines on  Cosal2015, CoSOD3k, and CoCA datasets. As shown in Table~\ref{Table:baselines}, the columns represent the baseline models and our method. The top two values for each metric are highlighted, with \colorbox{tab_red}{red} indicating rank-1 and \colorbox{tab_yellow}{yellow} indicating rank-2. For the Cosal2015 dataset, our method achieves five top-2 rankings. On the CoSOD3k dataset, our method secures six top-2 rankings. For the CoCA dataset, our method attains rank-1 across all six metrics. Overall, our method demonstrates the best performance across all three datasets.

\begin{table*}[tbp]
\caption{Co-saliency detection performance on clean datasets.
}
\centering
\label{Table:baselines}
\resizebox{0.8\linewidth}{!}{
\begin{tabular}{l|cccccc}
\hline 
\multirow{2}{*}{} & \multicolumn{6}{c}{Cosal2015}\tabularnewline
 & SR$\uparrow$  & IoU$\uparrow$ & MAE$\downarrow$  & maxF$\uparrow$ & $E^{max}_\xi$$\uparrow$ & $S_m$$\uparrow$ \tabularnewline
\hline 
GCAGC  & 0.8029 & 0.7162 & 0.0686 & 0.8139 & 0.8652 & 0.8137\tabularnewline
GICD  & 0.8382  & 0.7484 & 0.0744 & 0.8261 & 0.8774 & 0.8314\tabularnewline
GCoNet  & 0.8272  & 0.7472  & 0.0708 & 0.8327 & 0.8808 & 0.8324\tabularnewline
DCFM & 0.8486  & 0.7384  & 0.0678 & \cellcolor{tab_yellow}0.8446 & 0.8883 & 0.8291\tabularnewline
CADC   & \cellcolor{tab_yellow}0.8645 & \cellcolor{tab_red}0.7723 & \cellcolor{tab_yellow}0.0562 & \cellcolor{tab_red}0.8452 & \cellcolor{tab_yellow}0.8943 & \cellcolor{tab_yellow}0.8467\tabularnewline
\hline 
ConceptCoSOD  & \cellcolor{tab_red}0.8883 & \cellcolor{tab_yellow}0.7701 & \cellcolor{tab_red}0.0534 & 0.8412 & \cellcolor{tab_red}0.9041 & \cellcolor{tab_red}0.8467\tabularnewline
\hline 
\hline 
\multirow{2}{*}{} & \multicolumn{6}{c}{CoSOD3k}\tabularnewline
 & SR$\uparrow$  & IoU$\uparrow$ & MAE$\downarrow$  & maxF$\uparrow$ & $E^{max}_\xi$$\uparrow$ & $S_m$$\uparrow$ \tabularnewline
\hline 
GCAGC   & 0.7026 & 0.6260 & 0.0797 & 0.7362 & 0.8180 & 0.7674\tabularnewline
GICD  & 0.7361  & 0.6564  & 0.0883 & 0.7461 & 0.8326 & 0.7779\tabularnewline
GCoNet  & 0.7322  & 0.6584  & 0.0773 & 0.7569 & 0.8464 & 0.7847\tabularnewline
DCFM  & \cellcolor{tab_yellow}0.7677  & 0.6706  & \cellcolor{tab_red}0.0712 & \cellcolor{tab_red}0.7853 & \cellcolor{tab_yellow}0.8645 & \cellcolor{tab_yellow}0.7949\tabularnewline
CADC   & 0.7457 & \cellcolor{tab_yellow}0.6737 & 0.0816 & 0.7559 & 0.8414 & 0.7889\tabularnewline
\hline 
ConceptCoSOD  & \cellcolor{tab_red}0.8000 & \cellcolor{tab_red}0.6981 & \cellcolor{tab_yellow}0.0718 & \cellcolor{tab_yellow}0.7837 & \cellcolor{tab_red}0.8664 & \cellcolor{tab_red}0.8051\tabularnewline
\hline 
\hline 
\multirow{2}{*}{} & \multicolumn{6}{c}{CoCA}\tabularnewline
 & SR$\uparrow$  & IoU$\uparrow$ & MAE$\downarrow$  & maxF$\uparrow$ & $E^{max}_\xi$$\uparrow$ & $S_m$$\uparrow$ \tabularnewline
\hline 
GCAGC   & 0.3915 & 0.3766 & 0.0978 & 0.4724 & 0.6878 & 0.6360\tabularnewline
GICD  & 0.3683  & 0.4031  & 0.1455 & 0.4757 & 0.6778 & 0.6311\tabularnewline
GCoNet  & 0.3984  & 0.4206  & 0.1147 & 0.5089 & 0.7306 & 0.6517\tabularnewline
DCFM & \cellcolor{tab_yellow}0.4826  & \cellcolor{tab_yellow}0.4697  & \cellcolor{tab_yellow}0.0901 & \cellcolor{tab_yellow}0.5697 & \cellcolor{tab_yellow}0.7646 & \cellcolor{tab_yellow}0.6914\tabularnewline
CADC   & 0.4239 & 0.4406 & 0.1275 & 0.5034 & 0.6970 & 0.6569\tabularnewline
\hline 
ConceptCoSOD  & \cellcolor{tab_red}0.5652 & \cellcolor{tab_red}0.5290 & \cellcolor{tab_red}0.0886 & \cellcolor{tab_red}0.6091 & \cellcolor{tab_red}0.7698 & \cellcolor{tab_red}0.7201\tabularnewline
\hline 
\end{tabular}
}
\vspace{-20pt}
\end{table*}
\subsection{Comparison on Corrupted Datasets}
The robustness of our co-salient object detection method is essential, as real-world images often include various corruptions. To assess robustness, we applied common corruptions and an adversarial attack, as presented in Table~\ref{Table:baselines_corruption}. Specifically, we evaluated a weather-related corruption (frost), common distortions (motion blur, defocus blur, Gaussian noise) \cite{hendrycks2019robustness}, and an adversarial attack (Jadena) \cite{gao2022can}. The top two scores for each metric are highlighted, with \colorbox{tab_red}{red} indicating the highest and \colorbox{tab_yellow}{yellow} indicating the second highest. Results show that our method achieves top performance across all investigated corruption scenarios, including frost, motion blur, defocus blur, Gaussian noise, and adversarial attacks. Our method demonstrates superior robustness across a variety of corruptions.

\begin{figure}
\centering   
	\includegraphics[width=1\columnwidth]{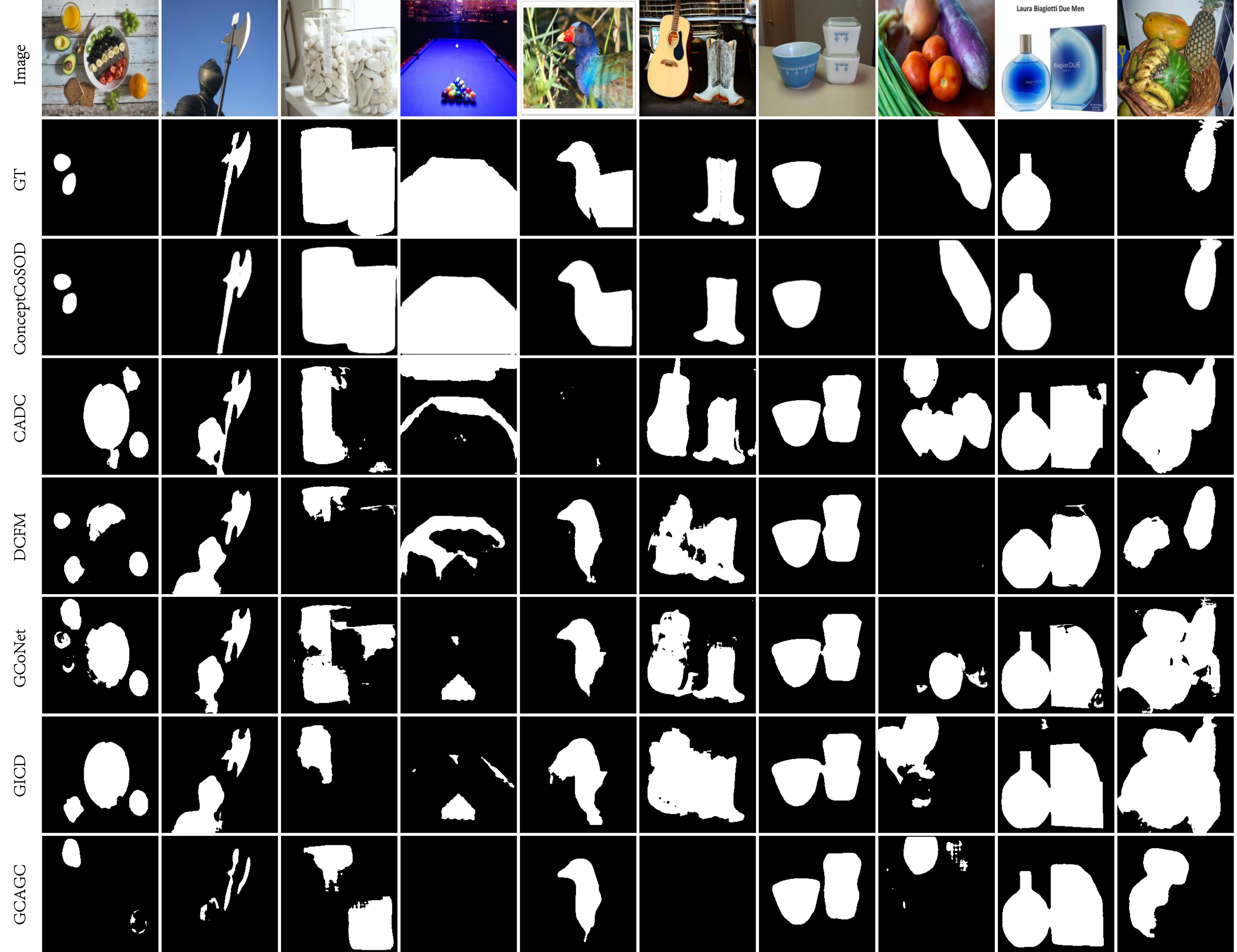}
	\caption{Visualization of our ConceptCoSOD method and other baselines on clean image dataset.
	}
	\label{fig:main_performance_ConceptCoSOD_clean}
\vspace{-20pt}
\end{figure}
\subsection{Visualization}
As shown in Fig.~\ref{fig:main_performance_ConceptCoSOD_clean}, we demonstrate the visualization result comparison between our method and baselines. Our method demonstrates precise segmentation of the target object, closely resembling the ground truth. In contrast, the baseline results are less accurate, often exhibiting incomplete segmentation of the target object or including non-target objects within the segmentation area.

\begin{figure*}[htbp]
	\centering   
	\includegraphics[width=1\linewidth]{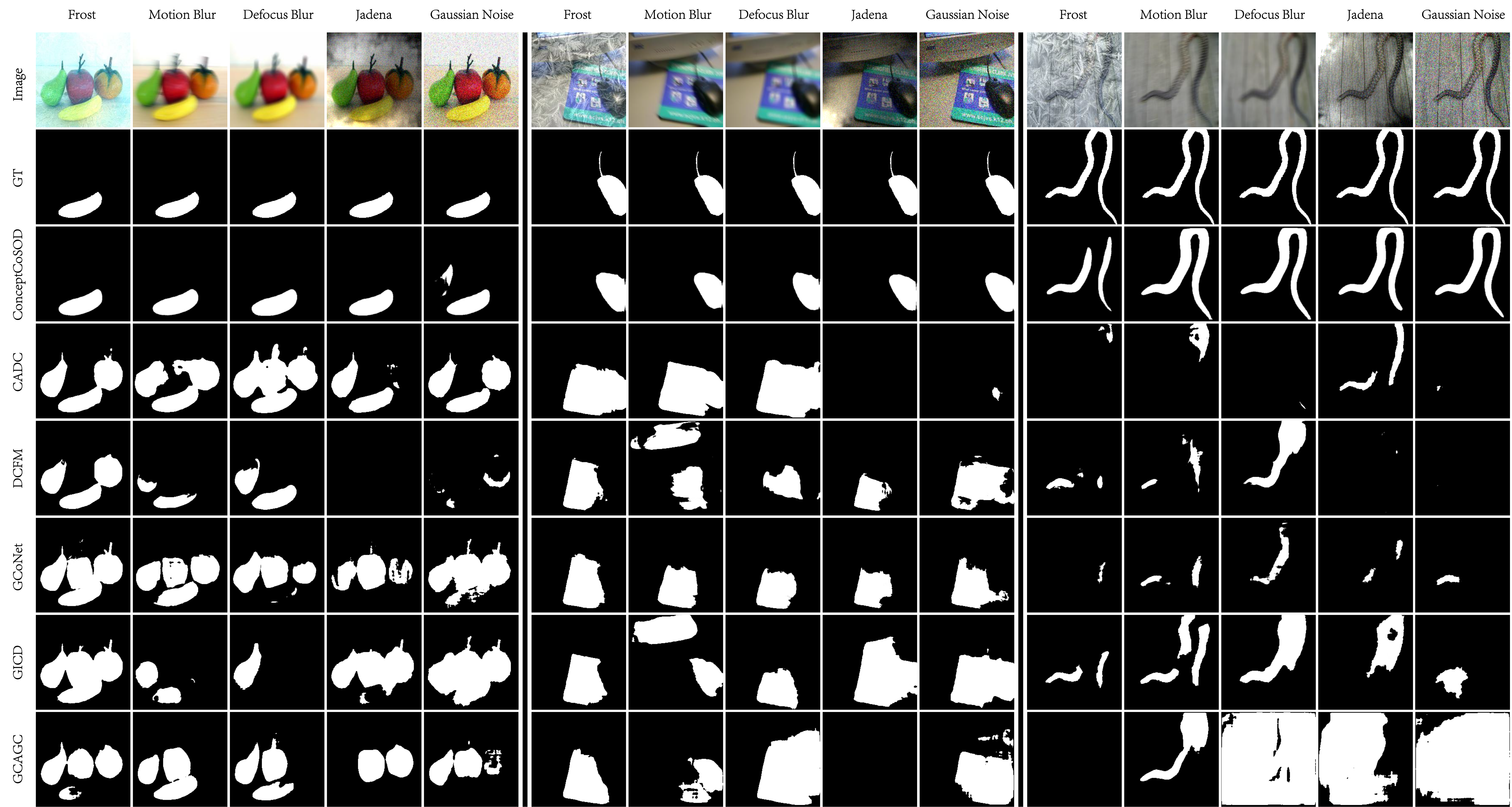}
	\caption{Visualization of our ConceptCoSOD method and other baselines on corrupted datasets.
	}
	\label{fig:main_performance_ConceptCoSOD_degra}
\end{figure*}

As shown in Fig.~\ref{fig:main_performance_ConceptCoSOD_degra}, we demonstrate the visualization result comparison between our method and baselines under corruptions. There are three cases, each containing five types of corruption. In most instances, only our ConceptCoSOD method effectively segments the target object, while other methods largely fail to identify it. This outcome demonstrates the superior robustness of our method against corruption compared to others.


\begin{table*}[tbp]
\caption{Co-saliency detection performance on corrupted datasets.
}
\centering
\label{Table:baselines_corruption}
\setlength{\tabcolsep}{8pt}
\resizebox{0.8\linewidth}{!}{
\begin{tabular}{l|cccccc}
\hline 
\multirow{2}{*}{} & \multicolumn{6}{c}{Frost}\tabularnewline
 & SR$\uparrow$  & IoU$\uparrow$ & MAE$\downarrow$  & maxF$\uparrow$ & $E^{max}_\xi$$\uparrow$ & $S_m$$\uparrow$\tabularnewline
\hline 
GCAGC & 0.6714 & 0.6114 & 0.0984 & 0.7233 & 0.7938 & 0.7489\tabularnewline
GICD & 0.7315  & 0.6589  & 0.1008 & 0.7581 & 0.8256 & 0.7745\tabularnewline
GCoNet & 0.6823  & 0.6196  & 0.1052 & 0.7380 & 0.8049 & 0.7539\tabularnewline
DCFM & 0.6848  & 0.6093  & 0.1037 & 0.7462 & 0.8056 & 0.7482\tabularnewline
CADC & \cellcolor{tab_yellow}0.7429 & \cellcolor{tab_yellow}0.6683 & \cellcolor{tab_yellow}0.0841 & \cellcolor{tab_yellow}0.7698 & \cellcolor{tab_yellow}0.8275 & \cellcolor{tab_yellow}0.7830\tabularnewline
\hline 
ConceptCoSOD & \cellcolor{tab_red}0.8198 & \cellcolor{tab_red}0.7130 & \cellcolor{tab_red}0.0809 & \cellcolor{tab_red}0.7929 & \cellcolor{tab_red}0.8567 & \cellcolor{tab_red}0.8055\tabularnewline
\hline 
\hline 
\multirow{2}{*}{} & \multicolumn{6}{c}{Motion Blur}\tabularnewline
 & SR$\uparrow$  & IoU$\uparrow$ & MAE$\downarrow$  & maxF$\uparrow$ & $E^{max}_\xi$$\uparrow$ & $S_m$$\uparrow$\tabularnewline
\hline 
GCAGC & 0.6853 & 0.6113 & 0.1063 & 0.7160 & 0.7935 & 0.7409\tabularnewline
GICD & 0.7052  & 0.6332  & 0.1088 & 0.7290 & 0.8117 & 0.7567\tabularnewline
GCoNet & 0.6203  & 0.5678  & 0.1105 & 0.6935 & 0.7683 & 0.7247\tabularnewline
DCFM & 0.6655  & 0.5948  & 0.0997 & 0.7287 & 0.7969 & 0.7404\tabularnewline
CADC & \cellcolor{tab_yellow}0.7260 & \cellcolor{tab_yellow}0.6504 & \cellcolor{tab_yellow}0.0907 & \cellcolor{tab_yellow}0.7431 & \cellcolor{tab_yellow}0.8212 & \cellcolor{tab_yellow}0.7686\tabularnewline
\hline 
ConceptCoSOD & \cellcolor{tab_red}0.8178 & \cellcolor{tab_red}0.6943 & \cellcolor{tab_red}0.0824 & \cellcolor{tab_red}0.7718 & \cellcolor{tab_red}0.8519 & \cellcolor{tab_red}0.7947\tabularnewline
\hline 
\hline 
\multirow{2}{*}{} & \multicolumn{6}{c}{Defocus Blur}\tabularnewline
 & SR$\uparrow$  & IoU$\uparrow$ & MAE$\downarrow$  & maxF$\uparrow$ & $E^{max}_\xi$$\uparrow$ & $S_m$$\uparrow$\tabularnewline
\hline 
GCAGC & 0.7057 & 0.6269 & 0.1090 & 0.7317 & 0.8039 & 0.7483\tabularnewline
GICD & \cellcolor{tab_yellow}0.7126  & \cellcolor{tab_yellow}0.6310  & 0.1059 & 0.7317 & \cellcolor{tab_yellow}0.8090 & \cellcolor{tab_yellow}0.7571\tabularnewline
GCoNet & 0.5781  & 0.5320  & 0.1180 & 0.6578 & 0.7320 & 0.7038\tabularnewline
DCFM & 0.6684  & 0.5958 & 0.0995 & \cellcolor{tab_yellow}0.7352 & 0.7968 & 0.7422\tabularnewline
CADC & 0.6928 & 0.6162 & \cellcolor{tab_yellow}0.0940 & 0.7121 & 0.7913 & 0.7490\tabularnewline
\hline 
ConceptCoSOD & \cellcolor{tab_red}0.7786 & \cellcolor{tab_red}0.6689 & \cellcolor{tab_red}0.0888 & \cellcolor{tab_red}0.7461 & \cellcolor{tab_red}0.8313 & \cellcolor{tab_red}0.7761\tabularnewline
\hline 
\hline 
\multirow{2}{*}{} & \multicolumn{6}{c}{Jadena}\tabularnewline
 & SR$\uparrow$  & IoU$\uparrow$ & MAE$\downarrow$  & maxF$\uparrow$ & $E^{max}_\xi$$\uparrow$ & $S_m$$\uparrow$\tabularnewline
\hline 
GCAGC & 0.5980 & 0.5501 & 0.1256 & 0.6502 & 0.7336 & 0.7025\tabularnewline
GICD & 0.6794  & 0.6162  & 0.1110 & 0.7111 & 0.7905 & 0.7481\tabularnewline
GCoNet & 0.5627  & 0.5257  & 0.1178 & 0.6447 & 0.7218 & 0.7014\tabularnewline
DCFM & 0.5995  & 0.5404  & 0.1097 & 0.6678 & 0.7412 & 0.7112\tabularnewline
CADC & \cellcolor{tab_yellow}0.7176 & \cellcolor{tab_yellow}0.6417 & \cellcolor{tab_red}0.0866 & \cellcolor{tab_yellow}0.7380 & \cellcolor{tab_yellow}0.8027 & \cellcolor{tab_yellow}0.7688\tabularnewline
\hline 
ConceptCoSOD & \cellcolor{tab_red}0.8049 & \cellcolor{tab_red}0.6996 & \cellcolor{tab_yellow}0.0883 & \cellcolor{tab_red}0.7801 & \cellcolor{tab_red}0.8465 & \cellcolor{tab_red}0.7950\tabularnewline
\hline 
\hline 
\multirow{2}{*}{} & \multicolumn{6}{c}{Gaussian Noise}\tabularnewline
 & SR$\uparrow$  & IoU$\uparrow$ & MAE$\downarrow$  & maxF$\uparrow$ & $E^{max}_\xi$$\uparrow$ & $S_m$$\uparrow$\tabularnewline
\hline 
GCAGC & 0.5320 & 0.5156 & 0.2074 & 0.6090 & 0.6666 & 0.6383\tabularnewline
GICD & 0.6853  & 0.6288  & 0.1224 & 0.7200 & 0.7995 & 0.7470\tabularnewline
GCoNet & 0.6054  & 0.5571 & 0.1183 & 0.6887 & 0.7663 & 0.7155\tabularnewline
DCFM & 0.6312  & 0.5659  & 0.1112 & 0.7044 & 0.7760 & 0.7217\tabularnewline
CADC & \cellcolor{tab_yellow}0.7141 & \cellcolor{tab_yellow}0.6457 & \cellcolor{tab_yellow}0.0949 & \cellcolor{tab_yellow}0.7461 & \cellcolor{tab_yellow}0.8149 & \cellcolor{tab_yellow}0.7652\tabularnewline
\hline 
ConceptCoSOD & \cellcolor{tab_red}0.7821 & \cellcolor{tab_red}0.6857 & \cellcolor{tab_red}0.0912 & \cellcolor{tab_red}0.7666 & \cellcolor{tab_red}0.8340 & \cellcolor{tab_red}0.7873\tabularnewline
\hline 
\end{tabular}
}
\end{table*}
\subsection{Ablation Study}
\noindent\textbf{Ratio of timestep resampling.}
Given the validation in Sec.~\ref{subsec:concept_learn} that performing denser sampling on the timesteps within the middle intervals is necessary, we further investigate the impact of different sampling ratios on the experimental results. Specifically, we examine the ratios of 1.25, 1.5, 1.75, and 2.0, and compare them with the base version (\ie, performing uniform sampling in all timesteps).
As shown in Table~\ref{Table:ratio_ablation}, all the validated ratio settings outperform the base version in performance as a whole. Among them, the 1.5 ratio yields the best results in the all six metrics. Therefore, we select 1.5 as the default resampling ratio for our experiments.

\begin{table}[tbp]
\caption{Results of our ConceptCoSOD method under different ratios of timestep resampling.
}
\centering
\label{Table:ratio_ablation}
\setlength{\tabcolsep}{8pt}
\resizebox{0.8\linewidth}{!}{

\begin{tabular}{l|cccccc}
\hline 
\multirow{2}{*}{} & \multicolumn{6}{c}{Cosal2015}\tabularnewline
 & SR$\uparrow$  & IoU$\uparrow$ & MAE$\downarrow$  & maxF$\uparrow$ & $E^{max}_\xi$$\uparrow$ & $S_m$$\uparrow$ \tabularnewline
\hline 
base & 0.8605 & 0.7507 & 0.0576 & 0.8236 & 0.8870 & 0.8358\tabularnewline
ratio 1.25 & 0.8694 & 0.7521 & 0.0593 & 0.8263 & 0.8897 & 0.8353\tabularnewline
ratio 1.5 & \cellcolor{tab_red}0.8883 & \cellcolor{tab_red}0.7701 & \cellcolor{tab_red}0.0534 & \cellcolor{tab_red}0.8412 & \cellcolor{tab_red}0.9041 & \cellcolor{tab_red}0.8467\tabularnewline
ratio 1.75 & 0.8689 & 0.7518 & 0.0609 & 0.8235 & 0.8887 & 0.8337\tabularnewline
ratio 2.0 & 0.8724 & 0.7585 & 0.0563 & 0.8317 & 0.8927 & 0.8406\tabularnewline
\hline 
\end{tabular}
}
\vspace{-5pt}
\end{table}

\noindent\textbf{Threshold.} In the Co-SOD task, the threshold for generating the final binary map is a key hyperparameter. To assess its influence, as shown in Table~\ref{Table:baselines_threshold}, we evaluate the results across a threshold range of [0.3, 0.8]. The dataset names are listed in the first column, with metrics in the second column and threshold values in the second row. Results indicate that threshold values of 0.5 provide optimal performance across six metrics. Therefore, we select 0.5 as the default threshold value for our experiments.
\begin{table}[tbp]
    \centering
	\caption{Results of our ConceptCoSOD method under different thresholds.}
	\label{Table:baselines_threshold}
	\setlength{\tabcolsep}{8pt}
	\resizebox{0.8\linewidth}{!}{
		\begin{tabular}{ll|cccccc}
\hline 
\multirow{2}{*}{} & \multirow{2}{*}{} & \multicolumn{6}{c}{threshold}\tabularnewline
 &  & 0.3 & 0.4 & 0.5 & 0.6 & 0.7 & 0.8\tabularnewline
\hline 
\multirow{6}{*}{\rotatebox{90}{Cosal2015}} & SR$\uparrow$  & 0.8526  & 0.8813  & \cellcolor{tab_red}0.8883 & 0.8669  & 0.8307  & 0.7483\tabularnewline
 & IoU$\uparrow$ & 0.7074 & 0.7466 & \cellcolor{tab_red}0.7701 & 0.7671 & 0.7338 & 0.6511\tabularnewline
 & MAE$\downarrow$ & 0.0821 & 0.0639 & 0.0534 & \cellcolor{tab_red}0.0506 & 0.0549 & 0.0688\tabularnewline
 & maxF$\uparrow$ & 0.7794 & 0.8171 & 0.8412 & \cellcolor{tab_red}0.8445 & 0.8223 & 0.7703\tabularnewline
 & $E^{max}_\xi$$\uparrow$ & 0.8702 & 0.8954 & \cellcolor{tab_red}0.9041 & 0.8955 & 0.8725 & 0.8140\tabularnewline
 & $S_m$$\uparrow$ & 0.7978 & 0.8281 & 0.8467 & \cellcolor{tab_red}0.8485 & 0.8322 & 0.7874\tabularnewline
\hline 
\hline 
\multirow{6}{*}{\rotatebox{90}{CoSOD3k}} & SR$\uparrow$  & 0.7545 & 0.7892 & \cellcolor{tab_red}0.8000 & 0.7882 & 0.7427 & 0.6444\tabularnewline
 & IoU$\uparrow$ & 0.6464 & 0.6821 & \cellcolor{tab_red}0.6981 & 0.6916 & 0.6539 & 0.5667\tabularnewline
 & MAE$\downarrow$ & 0.1008 & 0.0821 & 0.0718 & \cellcolor{tab_red}0.0679 & 0.0707 & 0.0825\tabularnewline
 & maxF$\uparrow$ & 0.7250 & 0.7637 & \cellcolor{tab_red}0.7837 & 0.7822 & 0.7600 & 0.7033\tabularnewline
 & $E^{max}_\xi$$\uparrow$ & 0.8336 & 0.8594 & \cellcolor{tab_red}0.8664 & 0.8568 & 0.8330 & 0.7701\tabularnewline
 & $S_m$$\uparrow$ & 0.7606 & 0.7897 & 0.8051 & \cellcolor{tab_red}0.8061 & 0.7888 & 0.7430\tabularnewline
\hline 
\hline 
\multirow{6}{*}{\rotatebox{90}{CoCA}} & SR$\uparrow$  & 0.5088 & 0.5613 & \cellcolor{tab_red}0.5652 & 0.5629 & 0.5328 & 0.4447\tabularnewline
 & IoU$\uparrow$ & 0.4864 & 0.5195 & \cellcolor{tab_red}0.5290 & 0.5128 & 0.4727 & 0.3916\tabularnewline
 & MAE$\downarrow$ & 0.1265 & 0.1037 & 0.0886 & 0.0789 & \cellcolor{tab_red}0.0731 & 0.0737\tabularnewline
 & maxF$\uparrow$ & 0.5605 & 0.5969 & \cellcolor{tab_red}0.6091 & 0.5951 & 0.5652 & 0.5063\tabularnewline
 & $E^{max}_\xi$$\uparrow$ & 0.7331 & 0.7677 & \cellcolor{tab_red}0.7698 & 0.7536 & 0.7234 & 0.6583\tabularnewline
 & $S_m$$\uparrow$ & 0.6753 & 0.7053 & \cellcolor{tab_red}0.7201 & 0.7178 & 0.7018 & 0.6616\tabularnewline
\hline 
\end{tabular}
	}
\vspace{-5pt}
\end{table}


\subsection{Limitation}
Our method performs well in general but remains sensitive to challenging cases. We plan to address this by incorporating advanced data augmentation strategies.

\subsection{Border Impact}
Our method enhances co-salient object detection by integrating semantic guidance, improving robustness in tasks like autonomous driving, medical imaging, and video analysis. This promotes more consistent and interpretable visual recognition in complex scenes. However, such improvements could be misused for surveillance or amplify biases from pre-trained models. We encourage ethical deployment, including fairness evaluation and human oversight, to ensure responsible use.

\section{Conclusion}
This paper presents a novel approach, ConceptCoSOD, which leverages text-based semantic guidance for co-salient object detection (Co-SOD). Unlike traditional Co-SOD methods that rely solely on image-based features, ConceptCoSOD introduces a robust (image-text)-to-image framework, enhancing object segmentation accuracy by incorporating semantic information from text. This approach showcases the potential of integrating text-driven semantics into Co-SOD tasks, providing a new direction for future research in vision-language models for complex segmentation tasks. In future work, we aim to enhance the robustness of concept learning to improve the method's practicality.

\clearpage
\newpage

{
    \small
    \bibliographystyle{neurips_custom}
    \bibliography{ref}
}

\end{document}